\documentclass[letterpaper, 10 pt, conference]{ieeeconf}  %

\IEEEoverridecommandlockouts                              %

\usepackage{graphics} %
\usepackage{times} %
\usepackage{amsmath} %
\usepackage{amssymb}  %
\usepackage{amsfonts}
\usepackage{subcaption}
\usepackage{hyperref}
\usepackage{graphicx}
\usepackage{float}
\usepackage{caption}
\usepackage{multirow}
\usepackage{color}
\usepackage{pifont}%
\usepackage{algorithm}
\usepackage{algpseudocode}
\algdef{SE}[DOWHILE]{Do}{doWhile}{\algorithmicdo}[1]{\algorithmicwhile\ #1}%

\title{\LARGE \bf
Intersection-free Robot Manipulation with Soft-Rigid Coupled Incremental Potential Contact
}

\author{Wenxin Du$^{*1}$, Siqiong Yao$^{*3}$, Xinlei Wang$^{2}$, Yuhang Xu$^{1}$, Wenqiang Xu$^{1}$, Cewu Lu$^{1}$%
\thanks{* indicates equal contribution.}%
\thanks{$^{1}${\tt\small \{mnkmYuki, yaosiqiong, xuyuhangtmx,  vinjohn, lucewu\}@sjtu.edu.cn}. Wenxin Du, Yuhang Xu, and Wenqiang Xu are with the School of Electronic Information and Electrical Engineering, Shanghai Jiao Tong University, Shanghai, China. Siqiong Yao is with SJTU-Yale Joint Center of Biostatistics and Data Science, National Center for Translational Medicine, Shanghai Jiao Tong University. Cewu Lu is the corresponding author, a member of Qing Yuan Research Institute and MoE Key Lab of Artificial Intelligence, AI Institute, Shanghai Jiao Tong University, Shanghai, China.}%
\thanks{$^{2}$Xinlei Wang is with ZenusTech Inc.
        {\tt\small wxlwxl1993@zju.edu.cn}}%
}

\begin{document}

\maketitle
\thispagestyle{empty}
\pagestyle{empty}

\begin{abstract}
This paper presents a novel simulation platform, ZeMa, designed for robotic manipulation tasks concerning soft objects.
Such simulation ideally requires three properties: two-way soft-rigid coupling, intersection-free guarantees, and frictional contact modeling, with acceptable runtime suitable for deep and reinforcement learning tasks. Current simulators often satisfy only a subset of these needs, primarily focusing on distinct rigid-rigid or soft-soft interactions. The proposed ZeMa prioritizes physical accuracy and integrates the incremental potential contact method, offering unified dynamics simulation for both soft and rigid objects. It efficiently manages soft-rigid contact, operating 75x faster than baseline tools with similar methodologies like IPC-GraspSim. To demonstrate its applicability, we employ it for parallel grasp generation, penetrated grasp repair, and reinforcement learning for grasping, successfully transferring the trained RL policy to real-world scenarios.
More experiments and videos can be found in the supplementary materials and on the website: \url{https://sites.google.com/view/zema-ipc}.
\end{abstract}

\section{INTRODUCTION}

Simulation platforms play a crucial role in developing and validating algorithms for robotic manipulation tasks. With the development of soft object manipulation and soft robots, there has been a growing interest in the simulation of soft-soft and soft-rigid interactions. However, these interactions still present significant challenges.  An ideal simulator for soft object manipulation should have three properties for accurate physics calculation: two-way soft-rigid coupling, intersection-free guarantee, and frictional contact modeling. From the application side, the simulator's runtime should be sufficiently fast to facilitate data generation for deep learning and reinforcement learning tasks.

\begin{figure}[t!]
    \centering
    \includegraphics[width=1\linewidth]{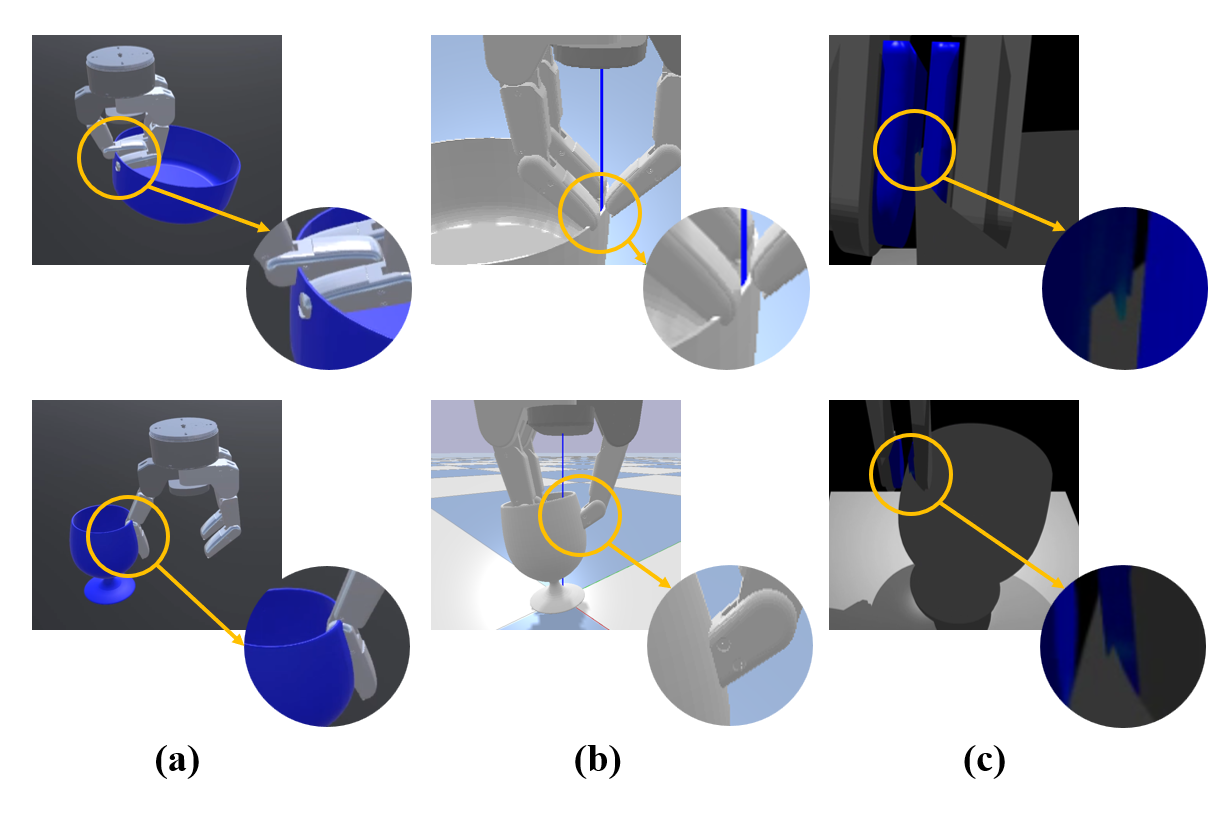}
    \caption{Model intersection occurs in common simulators: a. Isaac Gym, b. Unity, c. Pybullet even in rigid-rigid contact.}
    \label{fig:sim_penetration}
\end{figure}

Existing research on deformable object manipulation \cite{add,plasticinelab,rfuniverse,coupling_liu} often satisfies only a subset of these requirements. Regarding two-way coupling, most simulators typically focus on either rigid-rigid \cite{midas,nimble} or soft-soft \cite{diffcloth,flex,ipcgraspsim} contacts, employing independent solvers. Based on these simulators, a common soft-rigid coupling approach is to combine two distinct solvers using various bridging techniques \cite{coupling_liu}. Only a handful of methods \cite{plasticinelab,add} consolidate the soft-rigid interaction into a single solver. Regarding potential model intersections resulting from collisions, most simulators do not address them with precision. Often, these simulators represent the rigid object through simplified geometric primitives or convex hulls and detect collisions using approximated methods such as various discrete collision detection (DCD) techniques \cite{dcd1,dcd2}. Consequently, interpenetration issues, even in rigid-rigid interactions, can be observed in widely-used robot simulation platforms like Bullet \cite{pybullet}, Isaac Gym \cite{isaacgym}, and their derivative platforms \cite{orbit} (see Fig. \ref{fig:sim_penetration}). Regarding frictional contact modeling, a feature vital for robotic manipulation, but many simulators do not offer rigorous support. For example, the recent popular material point method (MPM) \cite{mpm} has been adopted for addressing soft-rigid coupling \cite{plasticinelab}, given its inherent capability to tackle the intersection problem with a reasonable runtime. Nonetheless, standard implementations of MPM simulators \cite{plasticinelab} often overlook frictional contact.

In this study, we introduce \textbf{ZeMa}, a simulator offering two-way soft-rigid coupling in a unified formulation, ensuring intersection-free and frictional contact. Drawing on the incremental potential contact (IPC) method \cite{ipc}, \textbf{ZeMa} uses unified IP energy to drive the dynamics simulation for both soft and rigid objects. In this framework, a soft object is represented by a tetrahedral mesh, while a rigid object is represented as an affine body \cite{abd}. Collisions between these object primitives are resolved using IPC. Though the coupling might seem straightforward by combining both object types' energies, the second-order solver of IPC necessitates intricate Hessian matrix calculations, further detailed in Sec. \ref{sec:unified_ip}. The entire system's implicit time integration is recast as a barrier-augmented, unconstrained nonlinear optimization problem. \textbf{ZeMa}'s robotic interface, based on the proposed unified IP energy, allows simulations at 3 fps for soft objects and 8 fps for rigid ones, marking an approximate 8x speed increase over IPC-GraspSim \cite{ipcgraspsim}. Owing to \textbf{ZeMa}'s GPU-parallelized interface, we can achieve more than 75x faster than IPC-GraspSim for data generation/sampling tasks.

Using \textbf{ZeMa}, we explore its utility across three applications: parallel grasp generation, penetrated grasp repair, and reinforcement learning for grasping. For the parallel grasp generation, we employ parallel techniques to produce multiple grasps concurrently. In the context of grasp dataset repair, we identify and rectify samples with penetrations from DexGraspNet \cite{dexgraspnet}, enhancing grasp quality. In the reinforcement learning (RL) task, we benchmark a grasping task involving rigid and soft cubes. We subsequently transfer the trained RL policy to a real-world setting. Besides, we conduct extensive ablation studies to demonstrate the stability and accuracy of \textbf{ZeMa}.

We summarize our contribution as follows:
\begin{itemize}
    \item We propose a unified incremental potential energy to address the two-way soft-rigid coupling in one solver, which can support intersection-free and frictional contact. Based on the solver, we present \textbf{ZeMa}, which contains a robotic interface for control and model training.
    \item We validate the usability of ZeMa in parallel grasp generation, penetrated grasp repair, and reinforcement learning for grasping. We also conduct extensive ablation studies to prove the simulator's physical accuracy and computational stability. 
\end{itemize}

\section{Related Works}
\label{sec:rel_works}
Our research presents a unified two-way soft-rigid coupling method applied to various domains. This section highlights the contact model and simulator application.

\subsection{Contact Simulation for Robot Manipulation}

Traditional analytical contact models, primarily static or quasi-static, don't provide the dynamic perspective offered by simulations that handle collision and model friction \cite{grasp_score,volume_metric,isotropy_metric}. Simulating contact with deformable objects, due to their extensive degrees of freedom and computational intensity, remains challenging.

Some approaches view contact as an impulse on a rigid surface and model it as a linear complementarity problem (LCP), as done by engines like ODE \cite{ode}, Bullet \cite{pybullet}, DART \cite{dart}, Drake \cite{drake}, and PhysX \cite{physx}. These methods approximate friction cones with a polyhedral structure. Conversely, MuJoCo approaches this as a convex optimization problem, but can sometimes yield non-complementary results \cite{mujoco}.
Compliant models \cite{add,compliant1,compliant2,brax,warp}, which consider deformable contact surfaces, eliminate the need for impulse calculations during collisions. However, deriving contact forces from these models requires unrealistic penetration.

Recently, the IPC method has gained traction in robotics \cite{ipcgraspsim, midas}. IPC-GraspSim \cite{ipcgraspsim} uses IPC to build a robotic grasping interface. Midas \cite{midas} operates on ABD and claims 15 fps performance with rigid-rigid interaction only. Our work diverges by enhancing the simulation process to accommodate affine and soft bodies, facilitating more efficient grasp generation and reinforcement learning training.

\subsection{Soft-rigid Coupling in Dynamics Simulation}
Soft-rigid coupling, pivotal for robotics involving deformable objects, brings challenges due to the differing representations and degrees of freedom between deformable and rigid entities. Conventionally, different objects are processed by distinct solvers and thus not coupled.

There are roughly three ways for soft-rigid coupling. The first kind is to treat the rigid object as a special form of soft object with high stiffness. In this way, the soft-rigid interaction can be uniformly treated by soft-soft dynamics solvers \cite{ipcgraspsim,plasticinelab}, although this increases the computational load on rigid objects.
The second kind is to utilize the existing individual solver as a black box, and try to bridge the force interaction between them \cite{coupling_liu}. DiffClothAI \cite{diffclothai} leverages Nimble \cite{nimble} and DiffCloth \cite{diffcloth} and implements two single-way coupling Cloth2Rigid and Rigid2Cloth. Bai and Liu \cite{coupling_liu} achieve two-way coupling between rigid objects and cloth via local patch augmentation.
The third kind is to unify the collision handling and frictional modeling. ADD \cite{add} and our work belong to this category since we both treat soft and rigid objects with different geometry representations but with the same contact modeling scheme.

\subsection{Applications of Robot Simulator}
The robot simulator is an important tool for algorithm development and can be applied to many scenarios. This work mainly demonstrates its potential for learning-based tasks on parallel grasp generation, dataset repairing, and reinforcement learning.

To generate grasp in simulators, most existing works \cite{grasp1,grasp2,grasp3} use the GraspIt! \cite{graspit} planner. However, since the planner searches in the low dimensional EigenGrasp space, the resulting data follows a narrow distribution and cannot cover the full dexterity of multi-finger hands. Recently, a large-scale dataset DexGraspNet \cite{dexgraspnet} is proposed. It generates grasps and evaluates the grasps in Isaac Gym \cite{isaacgym}.

Due to the recent development of robot learning, many simulation platforms exist for reinforcement learning tasks. However, since these simulation platforms are all based on the physics engines mentioned earlier, such as ODE, Bullet, Dart, they cannot guarantee intersection-free during contact. As shown in Fig. \ref{fig:sim_penetration}, even the recent popular Isaac Gym can encounter intersections when a hand grasps common objects like a bowl.
Since reinforcement learning requires considerable steps to optimize, IPC-GraspSim cannot meet the requirement as it takes $\sim 6$ seconds to process 1 step and cannot support parallel simulation.

\section{Unified Incremental Potential Energy} \label{sec:unified_ip}

Our work is built upon the incremental potential contact (IPC) method, we will first briefly describe it and then describe how we couple the soft and rigid in IPC.
The IP energy for a FEM-based soft body is:
\begin{equation}
    E(x, x^t, v^t)=\frac{1}{2}(x-\hat{x})^{T}M^{(x)}(x-\hat{x}) + h^2\Phi^{(x)}(x),
\end{equation}
where $x\in \mathbb{R}^{3N}$ is the soft body vertex position vector, $N$ is the number of vertices, $v=\dot{x}$ is the vertex velocity vector, $\hat{x}=x^t + hv^t + h^2M^{-1}f_e$, $f_e$ is the external forces applied on the vertices, $x^t$ and $v^t$ are the vertex positions and velocities at the last time-step respectively, $h$ is the time-step size, $M\in \mathbb{R}^{3N\times 3N}$ is the diagonal mass matrix for vertices, $\Phi(x)$ is the elastic potential energy for hyper-elastic materials.  

The IP energy for an affine rigid body \cite{abd} is:
\begin{equation}
\begin{split}
    E(q, q^t, \dot{q}^t) & = \sum_{b\in\mathcal{B}}(\frac{1}{2}(q_b - \hat{q}_b)^TM_{b}(q_b - \hat{q}_b) + h^2\Phi^{(q)}(q_b)) \\ 
    & = \frac{1}{2}(q-\hat{q})^TM^{(q)}(q-\hat{q}) + h^2\Phi^{(q)}(q),
\end{split}
    \label{eq:affine_IP}
\end{equation}
where $q_b=(p_b^T, a_1^T, a_2^T, a_3^T)^T\in \mathbb{R}^{12}$ is the state of an affine body $b\in\mathcal{B}$, $p$ and $A_b=(a_1, a_2, a_3)^T\in\mathbb{R}^{3\times3}$ are the associated translation vector and affine matrix. The transformation of the affine body vertex from its rest position $x^{(0)}$ to its current position $x$ is given by the mapping $x^{(0)} \mapsto x=A_bx^{(0)}+p_b$. We denote $J(x^{(0)})=\frac{\partial x}{\partial q_b} \in \mathbb{R}^{3\times 12}$. In Eq. \ref{eq:affine_IP},  $M_b=\int_{\Omega}\rho J(x^{(0)})^TJ(x^{(0)})dx^{(0)}$ is the mass matrix for affine body $b$, where $\Omega$ represents $b$'s material space; $\Phi^{(q)}(q_b)=\lambda V_b\lVert A_bA_b^T - I_3\rVert_F^2$ is the orthogonal energy characterizing the physical properties of affine bodies, where $\lambda$ is the stiffness parameter of affine bodies and $V_b$ is the volume of body $b$. By using large $\lambda$, affine bodies can be nearly rigid. Therefore, we can use affine bodies to model real-world rigid bodies. 

An intuitive way to integrate the two kinds of dynamic systems is to simply combine the two energy terms as in:
\begin{equation}
\begin{split}
E(x_s, q, x_s^t, q^t, v_s^t, \dot{q}^t) =& \frac{1}{2}(x_s-\hat{x}_s)^{T}M^{(x_s)}(x_s-\hat{x}_s) \\&+ \frac{1}{2}(q-\hat{q})^{T}M^{(q)}(q-\hat{q}) \\&+ h^2\Phi^{(x_s)}(x_s) + h^2\Phi^{(q)}(q), 
\end{split}
\label{eq:coupled-ip-energy}
\end{equation}
where the subscript $s$ stands for soft bodies, and $\hat{x}_s=x_s^t + hv_s^t + h^2(M^{(x_s)})^{-1}f_e^{(x_s)}$, $\hat{q}=q^t + h\dot{q}^t + h^2(M^{(q)})^{-1}f_e^{(q)}$.

However, the optimization for Eq. \ref{eq:coupled-ip-energy} cannot be directly done together as collision needs to be handled.
Therefore, we extend the barrier-augmented IP energy in IPC \cite{ipc} to :

\begin{equation}
\begin{split}
B_t(x_s,q) & = E(x_s, q, x_s^t, q^t, v_s^t, \dot{q}^t) + B(x_s, q)\\
& :=E(x_s, q, x_s^t, q^t, v_s^t, \dot{q}^t) +\kappa \sum_{k\in C}b(d_k(x_s,q)),
\end{split}
\end{equation}
where 
\begin{equation}
b(d) = b(d, \hat{d})=
\begin{cases}
  -(d - \hat{d})^2log(\frac{d}{\hat{d}}), &0<d<\hat{d}\\
  0 &d\geq \hat{d}
\end{cases}
\end{equation}
is the barrier function in IPC, $d_k$ is the distance between the contact pair $k$, $\hat{d}$ is a distance threshold parameter, and $C$ is the set of contact pairs between all affine and soft bodies.

Solving this soft-affine coupled system by Projected Newton requires barrier energy gradient and Hessian computation for collision pair involving primitives (vertices, edges, and triangles) of affine and soft body. 
For example, for a collision pair containing a soft body triangle $(t_0, t_1, t_2)$ and a vertex of an affine body $b$, we have: 

\begin{equation}
\begin{pmatrix} x_{s, t0}\\ x_{s, t1}\\ x_{s, t2} \\ x_p \end{pmatrix}=\begin{pmatrix} I_9 && O \\ O && J_{b}(x^{(0)}_p) \end{pmatrix}\begin{pmatrix} x_{s, t0}\\ x_{s, t1}\\ x_{s, t2} \\ q_b \end{pmatrix}
\end{equation}
We denote this equation by $\tilde{x}_{pt}=\tilde{J}\tilde{y}_{pt}$, and our target is to compute the gradient and hessian of $d_{pt}$ w.r.t. $\tilde{y}_{pt}$. By the chain rule, we have: 
\begin{equation}
\nabla_{\tilde{y}_{pt}}d_{pt}=\tilde{J}^T\nabla_{\tilde{x}_{pt}}d_{pt}=
\begin{pmatrix}
\nabla_{x^t}d_{pt}\\
J_b(x_p^{(0)})^T\nabla_{x_p}d_{pt}
\end{pmatrix}
\end{equation}

\begin{equation}
\frac{\partial^2 d_{pt}}{\partial \tilde{y}_{pt}^2}=\tilde{J}^T\frac{\partial^2d_{pt}}{\partial \tilde{x}_{pt}^2}\tilde{J}=
\begin{pmatrix}
H_{tt} & H_{tp}J \\
J^TH_{pt} & J^TH_{pp}J
\end{pmatrix},
\end{equation}
where $H_{pt}=\frac{\partial^2 d_{pt}}{\partial x_p \partial \tilde{x}_t}$ and $\tilde{x}_t=(x_{s, t0}, x_{s, t1}, x_{s, t2})^T$. $H_{pp}$, $H_{tt}$, and $H_{tp}$ have similar meanings. 

And we can subsequently project $\frac{\partial^2 d_{pt}}{\partial\tilde{y}_{pt}^2}$ into a semi-positive definite matrix (SPD) by projecting $\frac{\partial^2 d_{pt}}{\partial\tilde{x}_{pt}^2}$ into a SPD due to the above equation. In this way, we reduce the gradient and hessian computation of the soft-rigid coupling barrier energy into the common case in IPC. Edge-edge collision pairs also share a similar process to this point-triangle case. 

For friction forces, IPC adopts a variational approximation of the Coulomb friction model: 
\begin{equation}
f_k = -\mu\lambda_k f_1(\lVert u_k\rVert) T_k(x)\frac{u_k}{\lVert u_k\rVert} \\ 
\end{equation}
\begin{equation}
f_1(x)=
\begin{cases}
-\frac{y^2}{\epsilon_v^2h^2} + \frac{2y}{\epsilon_vh}, &y\in(0, h\epsilon_v)\\
1, & y\geq h\epsilon_v
\end{cases}, 
\end{equation}
where $f_k$ is the approximated friction force, $\epsilon_v$ is a velocity threshold parameter, $T_k(x)$ is the sliding basis of the contact pair $k$, $u_k=T_k(x)^T(x-x^t)$ is the tangential relative displacement within the contact pair $k$. 
Such a smooth approximation allows IPC to integrate friction force into the variational framework by constructing a frictional energy $D_k(x,x^t)=\mu\lambda_k^nf_0(\Vert u_k\Vert)$, where $\mu$ is the friction coefficient, $\lambda$ is the magnitude of the normal contact force given by the gradient of the barrier energy, $f_0$ is a function satisfying $\dot{f_0}=f_1$ and $f_0(\epsilon_v h)=\epsilon_v h$. Analogous to the barrier energy, we can properly handle the soft-rigid coupled friction $D_k(x_s, q, x_s^t, q^t)$. Adding the total friction energy $D(x_s, q, x_s^t, q^t)=\sum_{k\in C}D_k(x_s, q, x_s^t, q^t)$ to the barrier augmented incremental potential energy, we derive the total incremental potential contact (IPC) energy. Finally, we can solve the state at the next time step by the equation: 
\begin{equation}
x_s, q = argmin_{x_s, q} B_t(x_s, q, x_s^t, q^t, v^t, \dot{q^t}) + D(x_s, q, x_s^t, q^t) 
\end{equation}
The rigid body vertex positions can be obtained by $x_a = Jq$, and the velocity $v_s$ and $\dot{q}$ will be updated to $\frac{1}{h}(x_s-x_s^t)$ and $\frac{1}{h}(q-q^t)$ respectively.

\section{ZeMa}
To simulate the manipulation task with the unified energy, we implement a robotic manipulation interface, which supports both the 1-DoF parallel-jaw grippers and the high-DoF dexterous grippers.
Given a gripper with state $\theta=(\theta_{pose}, \theta_{joint})$, and object location $o$, the gripper will be driven to manipulate the object with a control scheme.

In this work, the manipulation task is grasp, and control is position control. Following \cite{rigidipc}, we add a quadratic energy term to drive the gripper links to the target position using the Augmented Lagrangian optimization algorithm.  

\subsection{Controller}\label{sec:controller}
ZeMa can import robot models in URDF or MJCF format and support two control methods for them: kinematic control and PD control. Since we adopt kinematic control for the main experiments, we leave the description of PD control in supplementary materials.

We first define a joint state increment $\Delta \theta_{joint}$ to indicate the gripper state transition from $\theta_{joint}$ to $\theta_{joint} + \Delta \theta_{joint}$. Then we designate the target state of the gripper at the next time step to be $\theta_{joint} + \phi(f)\Delta \theta_{joint}$ by adding the kinematic constraints which will be discussed in section \ref{sec:constraints}, where $f$ is the force magnitude measured at the gripper fingers given by the simulator. Details about the force measurement in ZeMa can be found in section \ref{sec:force-sensing}. Here, the function $\phi(f)$ is a function that decays exponentially before it reaches the desired force magnitude $f^{*}\geq 0$, which is as follows:
\begin{equation}
\phi(f)=
\begin{cases}
  (e^{-kf} - e^{-kf^*})/(1-e^{-kf^*}) &, 0\leq f<f^*\\
  0 & , f \geq f^*
\end{cases}
\end{equation}

In our experiments, we choose $k=20 N^{-1}$, $f^*=0.5N$.%

\subsection{Contact Force Sensing}
\label{sec:force-sensing}
Thanks to the potential energy formulation of contact forces in ABD and IPC, we can directly compute the vertex forces by computing the barrier energy gradient $f_{c}=\frac{\partial B(x, \hat{d})}{\partial x}$. The resultant contact force applied on a rigid body $b$ can be derived by simply adding all its vertex forces. 

\subsection{Constraints}
\label{sec:constraints}
To fully support articulated object simulation and robotics applications, such as repairing mesh intersections in contact-rich datasets, we integrate 3 kinds of constraints into ZeMa. 

\subsubsection{Kinematic Constraints}
ZeMa supports kinematic object modeling, %
so that any contact cannot affect the state of a kinematic object, but the kinematic object can affect the states of other objects through contact. It can achieve one-way rigid2soft coupling and can be used for soft object manipulation. Similar to Rigid-IPC \cite{rigidipc}, we implement the constraints by the Augmented Lagrangian algorithm introducing a kinematic energy term for an affine body $b$: 
\begin{equation}
E_A(q_b, \lambda_A)=\frac{\kappa_A}{2}\Vert q_b - \tilde{q_b}\Vert_2^2 - \sqrt{m_k}\lambda_A^T(q_b - \tilde{q_b})
\end{equation}
where $\tilde{q_b}$ is the target state of $b$, $\kappa_A$ denotes the penalty stiffness, and $\lambda_A$ is the Lagrangian multiplier. 

\subsubsection{Joint Constraints}
Similar to kinematic constraints, joint constraints for articulated objects are also implemented by applying Augmented Lagrangian algorithm. Our joint constraint formulation is mainly consistent to the non-linear constraint and inequality constraint formulation in \cite{ipcarticulate}. For a prismatic joint constraint between two affine bodies $b_1$ and $b_2$ , we use a different non-linear and simpler energy term $E_{prismatic}=\kappa_A(\Vert A_{b_1} - A_{b_2} \Vert_2^2 + \Vert((A_{b_1}c + p_{b_1}) - (A_{b_2}c + p_{b_2})) \times (A_{b_1}d)\Vert^2)$, where $d$ is the prismatic joint axis coordinates in the local frame of $b_1$, $c$ can be any point theoretically, here we choose it to be the rest position of $b_1$'s center of mass. Similar to kinematic constraints, we also add a lagrangian multiplier to force the constraint $\Vert A_{b_1} - A_{b_2} \Vert_2^2 + \Vert((A_{b_1}c + p_{b_1}) - (A_{b_2}c + p_{b_2})) \times (A_{b_1}d)\Vert^2 = 0$. 

\subsubsection{Spring Constraints}
In applications like penetrated dataset repair (Sec. \ref{exp:repair}), we usually need to find an intersection-free state nearest to a given target state. In such cases, simply linking the objects to the target position with springs can be extremely helpful. In ZeMa, spring constraints are supported by directly adding the spring energy $E_{spring}=k_{spring}\Vert x - x_{target} \Vert_2^2$ to the IP energy. Here $k_{spring}$ represents the spring stiffness coefficient.

\subsection{Constitutive Models}
ZeMa's modifiable parameters include the stiffness of rigid objects, Young's modulus, Poisson ratio of soft objects, garment stretching and shearing energies, and the friction coefficient. The constitutive model of soft solid bodies can be chosen within Saint Venant–Kirchhoff, fixed corotated, and neo-Hookean models. Saint Venant-Kircchoff model is a simple non-linear extension of the linear elastic model. The fixed corotated model is another simple rotational invariant model. The neo-Hookean energy model with logarithm terms could resist such compression since its non-linear energy increases to infinity when the volume becomes zero. ZeMa also supports the Baraff-Witkin model, which properly handles anisotropic stretch or compression and bending of garments \cite{largestep}. In our experiments, we choose the friction coefficient $\mu$ to be $0.4$ in most cases. We use the neo-Hookean model for soft bodies due to its stability under large deformation; for garments, we apply the Baraff-Witkin model. 

\begin{figure}[t!]
    \centering
    \includegraphics[width=1\linewidth]{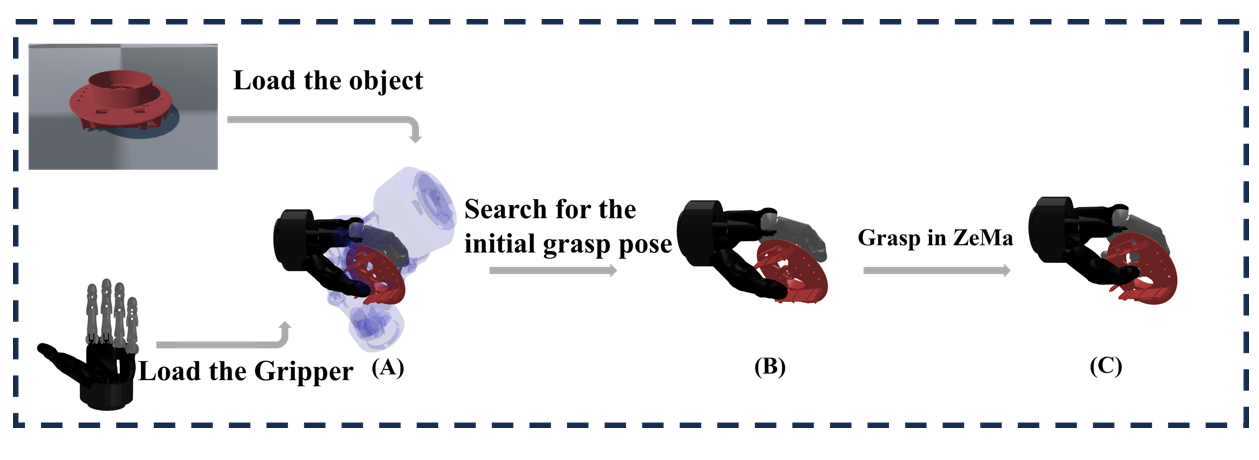}
    \caption{Overall procedure of robotics gripper grasping interface. We first load the object mesh and the robot gripper into the simulator, where a grasp planner (A) searches for an initial grasp pose (B). Then, we use ZeMa to perform the grasp (C).}
    \label{fig:grasp_process}
\end{figure}

\subsection{Simulation Procedure}\label{sec:sim_proc}
The overall procedures are illustrated in Fig. \ref{fig:grasp_process}.
\subsubsection{Grasp Initialization} To enable both parallel-jaw and dexterous grasping, we adopt a grasp planning algorithm, Iterative Surface Fitting (ISF) \cite{isf}, which is efficient and easy to parallel. We reimplement the ISF with GPU and denote it ``parallel ISF''. It can generate 25 grasps in 2.5 seconds with Shadow hand. 

\subsubsection{AutoGrasp}\label{sec:autograsp} After the initial pose generated, we use the kinematic control mode to close gripper fingers following the scheme previously introduced in Sec \ref{sec:controller}. We stop when the contact force for each finger reaches 1N. 

\subsubsection{Lift} Once the contact force reaches 1N. We lift the object from the ground by 5 cm to verify the planned grasp. 

\subsubsection{Shake} To ensure the stability of a grasp, we shake the object after lifting it. A successful stable grasp is determined when, at the end of the simulation, the object is still in contact with the hand and not in contact with the ground.

\section{Applications}
\subsection{Parallel Grasp Generation}\label{sec:parallel}
To validate the ability of grasp generation, we compare with IPC-GraspSim \cite{ipcgraspsim}. For the object set, we found 2 objects from the IPC-GraspSim object set because they are all available online to the best of our knowledge. Additionally, we select 9 objects from the adversarial dataset in Dex-Net \cite{dexnet}. The object samples can be referred to our website.

Then, we perform the grasp operation described in Sec. \ref{sec:sim_proc} in ZeMa with objects using the kinematic controller. Due to the GPU-based simulation pipeline, we can easily make the grasp operation process parallel. Each grasp operation occupies $\sim$170 MB of GPU memory on average, but an extra GPU memory with a fixed size of $\sim$7GB is always required for the parallel bounding volume hierarchy (BVH) cache and a sparse matrix cache. These caches reduce the need for frequent, time-consuming device memory allocation. On a GeForce RTX 4090 graphics card, we can evaluate 25 grasps in $\sim$120 seconds within 60 frames, while running a single grasp requires $\sim$44 seconds. As a result, ZeMa achieves a $\sim$10$\times$ acceleration on the grasp evaluation task by parallel computing. 
Table \ref{tab:breakdown_runtime} shows the breakdown of the consumed time of each component in a typical grasp simulation case. The discrete collision detection, implemented by BVH, occupies most of the runtime, as the parallel BVH suffers from poorly coalesced memory access. Energy computation runtime is lifted by the thread divergence issue introduced by various complex constraints-related logic branches. We leave further computation efficiency improvement in future works.

\begin{table}[ht!]
    \centering
    \begin{tabular}{|c|cccccc|}
    \hline
       Component  & DCD & Energy  & Hess & Linear & CCD & Others\\ \hline
       Runtime (s) & 17.06 & 6.04 &  1.72 & 0.3 & 0.005 & 1.65 \\ \hline
    \end{tabular}
    \caption{Breakdown analysis of components runtime. \textbf{DCD}, \textbf{Energy}, \textbf{Hess}, \textbf{Linear}, \textbf{CCD} and \textbf{Others} stands for the processes of Discrete Collision Detection, energy computation, Hessian matrix computation, linear system solving, Continuous Collision Detection, and other minor operations such as synchronizing data between CPU and GPU devices respectively.}
    \label{tab:breakdown_runtime}
\end{table}

\begin{figure}[h!]
    \centering
    \includegraphics[width=1\linewidth]{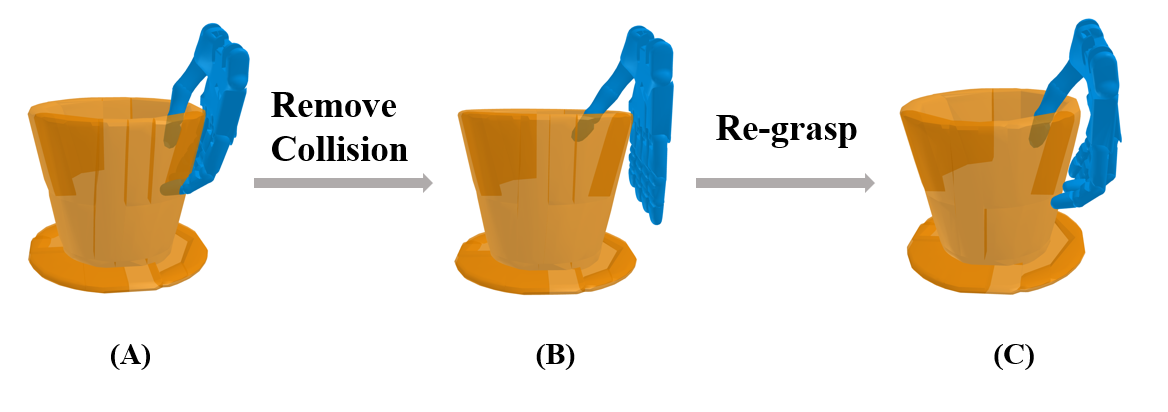}
    \caption{The procedure of repairing grasps in the DexGraspNet penetration subset. We remove the collision (B) from an interpenetrated grasp (A), and re-grasp in ZeMa (C). We set spring constraints linking the gripper links to their original positions in ZeMa to ensure a result close to the original pose (A).}
    \label{fig:fix_dexgraspnet_grasps}
\end{figure}
\subsection{Penetrated Grasp Repair}\label{sec:grasp_penetration_fix}
Beyond parallel grasp generation, ZeMa is adept at rectifying mesh intersections and the negative volume issues in existing datasets, as shown in Fig. \ref{fig:fix_dexgraspnet_grasps}. We selected 496 grasps from DexGraspNet, each exhibiting significant penetration originally. For each grasp, we expanded the fingers of the Shadow hand using its joint parameters and repositioned the gripper away from the object, along the opposite of the palm face normals, until intersections were eliminated. By introducing spring constraints to drive the gripper links to the original DexGraspNet poses, we perform AutoGrasp described in Sec. \ref{sec:autograsp} to obtain intersection-free grasps.

\subsection{Reinforcement Learning Benchmark}

To benchmark the reinforcement learning algorithm, we implement a gym-style interface, showcasing a cube-picking task using a Franka Emika panda robot. The cube may be rigid or deformable. ZeMa inherently manages the contact between the gripper and the cube, irrespective of whether it is rigid-rigid or rigid-soft. The task goal is to lift the object from the table. The task's reward, $R=R_{reach} + R_{pick} + R_{border}$, is formulated where $R_{reach}=-\Vert x_{gripper} - x_{object}\Vert_2^2$ guides the gripper towards the object. $R_{pick}$ is $5$ when the cube is successfully lifted, otherwise it's $0$. $R_{border}$ penalizes when the gripper exits the defined range with a value of -10. The action $a=(\Delta \vec{x}, \Delta \theta) \in \mathbb{R}^3$ consist of horizontal displacement and vertical axis rotation.
The observation $o=(s_{gripper}, s_{object}, s_{object} - s_{gripper})\in \mathbb{R}^9$ is a function of the gripper state $s_{gripper}$ and the object state $x_{object}$. The gripper state $s_{gripper}\in \mathbb{R}^3$ is a concatenation of the gripper x-coordinate, z-coordinate, and its y-rotation (y-axis is the vertical one). The object state $s_{object}$ has a similar meaning.  In each episode, the agent will move for $N=64$ steps and try to pick up the cube at the final step in a top-down fashion. We choose PPO \cite{ppo} as the RL training algorithm and train the agent five times for 15, 000 steps.

\section{Experimental Results}
\label{sec:result}
\subsection{Metrics}
\noindent\textbf{F1 score for Grasp} A harmonic mean of average precision (AP) and average recall (AR). It is inherited from IPC-GraspSim \cite{ipcgraspsim}.\\
\textbf{Penetration Depth for Grasp} The maximal penetration for a robot hand to the object mesh, the unit is centimeter.\\
\textbf{Success Rate} For the penetrated grasp repair task, it is the successful grasp number after repair over all the penetrated grasp. For the RL task, the successful grasp number over all the grasp trials.

\subsection{Contact Simulation Comparison}
We generate the top-down grasps by following the procedures in IPC-GraspSim with the object set described in Sec. \ref{sec:parallel}. These grasps are validated in IPC-GraspSim, Isaac Sim, and ZeMa. As shown in Table \ref{tab:grasp-sim}, we can observe a strong positive correlation between the F1 scores of IPC-GraspSim and ours, yet the runtime of ZeMa is far lower than Isaac Gym and IPC-GraspSim. 
\begin{table}[ht!]
    \centering
    \begin{tabular}{|c|cccc|}
    \hline
       Simulation  & AP & AR & F1  & Runtime (s)  \\\hline
       Isaac Gym  & 0.75 & 0.79 & 0.76  & 421 \\ \hline
       IPC-GraspSim & 0.84 & 0.79 & 0.83  & 605\\ \hline
       ZeMa & 0.79 & 0.77 & 0.80 &  72\\\hline
       ZeMa (Parallel) & 0.79 & 0.77 & 0.80 & 8 \\ \hline 
    \end{tabular}
    \caption{The contact simulation results of Isaac Gym, IPC-GraspSim, and ZeMa. ZeMa evaluates grasps 8 times faster than IPC-GraspSim. }
    \label{tab:grasp-sim}
\end{table}

IPC-GraspSim marginally outperforms ours because its gripper finger is modeled as a soft object, resulting in increased contact points and frictional forces. This advantage, however, comes at the expense of runtime efficiency and simulation parameter adjustments. The complexity of soft-soft contact dynamics often challenges simulation stability, and we've observed stability issues in IPC-GraspSim, which we detail in the supplementary materials. Nonetheless, given that our simulator surpasses the popular Isaac Gym in both performance and runtime, it can prove the efficacy of ZeMa.

\subsection{Application: Penetrated Grasp Repair}\label{exp:repair}
As shown in Table \ref{tab:fix_dexgraspnet}, the average penetration depth of the Dexgrapnet penetration subset previously described in sec. \ref{sec:grasp_penetration_fix} is 3.69 cm. By applying ZeMa, we obtain a penetration-free version of the subset. Detailed success rates and penetration depth data can be found in Table \ref{tab:fix_dexgraspnet}.  
\begin{table}[ht!]
    \vspace{0.2cm}
    \centering
    \begin{tabular}{|c|cc|}
    \hline
       Dataset  & success  & pen(cm)  \\ \hline
       DexGraspNet(subset) & 0\%  &  3.69  \\ \hline
       DexGraspNet-fixed(subset)  & 2.8\% & 0\\ \hline
    \end{tabular}
    \caption{The success rate and the average penetration depth of the DexGraspNet penetration subset data (DexGraspNet(subset)) and the data fixed by ZeMa (DexGraspNet-fixed(subset)). By repairing mesh intersections within the subset, we obtain grasps without penetration. }
    \label{tab:fix_dexgraspnet}
     \vspace{-0.2cm}
\end{table}

The reason why the dataset after repair still has a low success rate is that most grasp poses provided by DexGraspNet are not valid grasps. The robot hand and the object are posed to make contact with each other but do not guarantee a grasp pose. We demonstrate some failure cases in the supplementary materials. 

\subsection{Application: Reinforcement Learning Benchmark}

\begin{figure}[t!]
\vspace{0.2cm}
\centering
\begin{subfigure}{0.48\linewidth}
    \includegraphics[width=\linewidth]{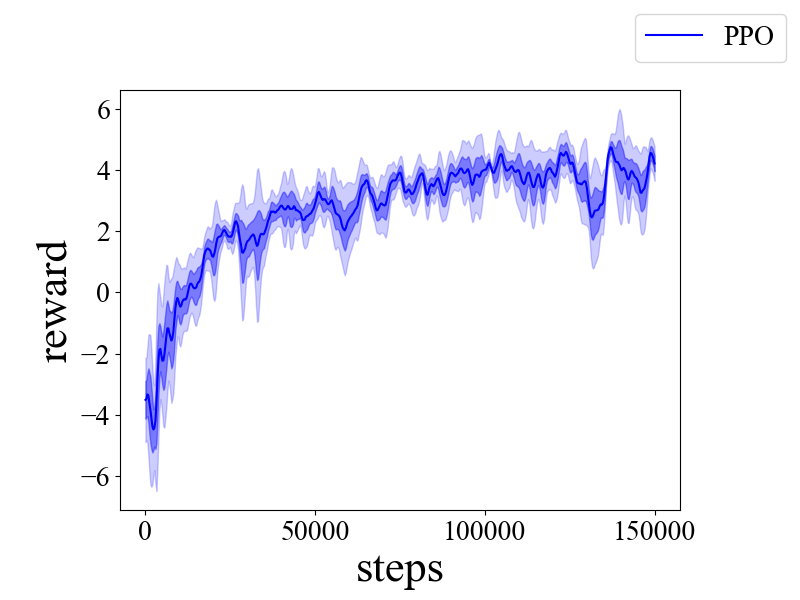}
    \label{fig:first}
\end{subfigure}
\hfill
\begin{subfigure}{0.48\linewidth}
    \includegraphics[width=\linewidth]{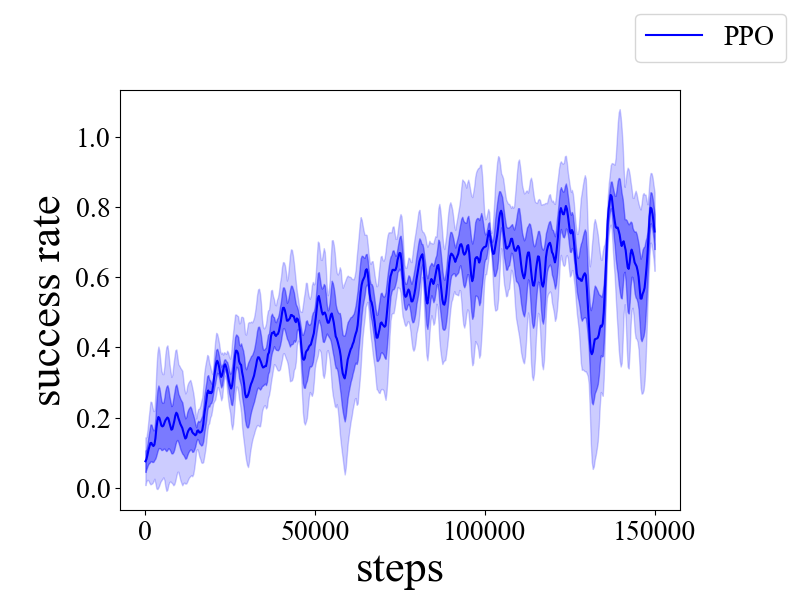}
    \label{fig:second}
\end{subfigure}
        
\caption{The reward and success rate of the cube-picking task during training.}
\label{fig:RL}
\vspace{-0.4cm}
\end{figure}

The agent can achieve an average success rate of 73\% after training in the simulator. Reward and success rate curves are illustrated in Fig. \ref{fig:RL}.  For real-world experiments, since the RL environment observation depends only on the gripper state $s_{gripper}$ and the object state $s_{object}$, we stick QR code markers onto the cube object surfaces to track its transformation. After aligning the real-world coordinate frame with the simulation one, we can directly feed the trained RL agent real-world observation to acquire the target gripper position and subsequently use PD control to move the gripper to that target in each step. We conducted 10-time real-world experiment, resulting in a success rate of 70\%. 

\subsection{Ablation Study}
\paragraph{Sensitivity of Soft Material Simulation}
We test the sensitivity of solid soft materials by simulating a solid soft cube with Young's modulus and Poisson's ratio falling onto the ground due to gravity. In addition, we drop a piece of garment with different Young's modulus onto a rigid cone to test the sensitivity of the garment material modeling. 

\begin{figure}
    \vspace{0.2cm}
    \centering
    \includegraphics[width=1\linewidth]{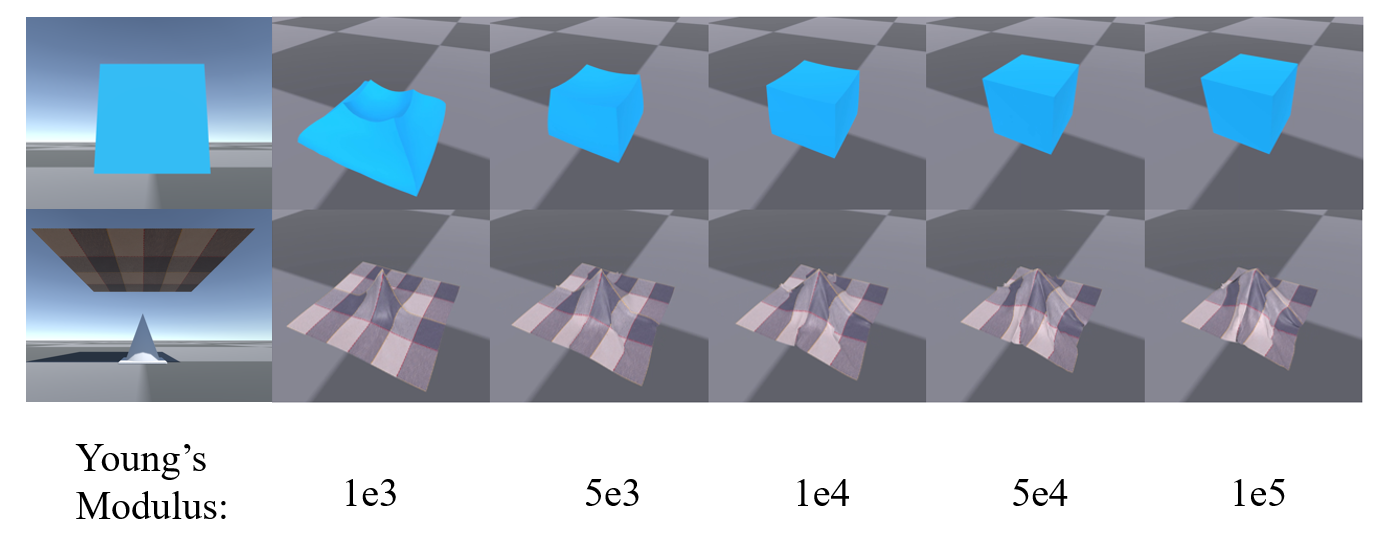}
    \caption{The first row shows five soft cubes with different Young's modulus falling on the ground. The second row includes garments with different Young's modulus falling onto a rigid cone. }
    \label{fig:ablation-youngs-modulus}
\end{figure}

\begin{figure}
    \centering
    \includegraphics[width=1\linewidth]{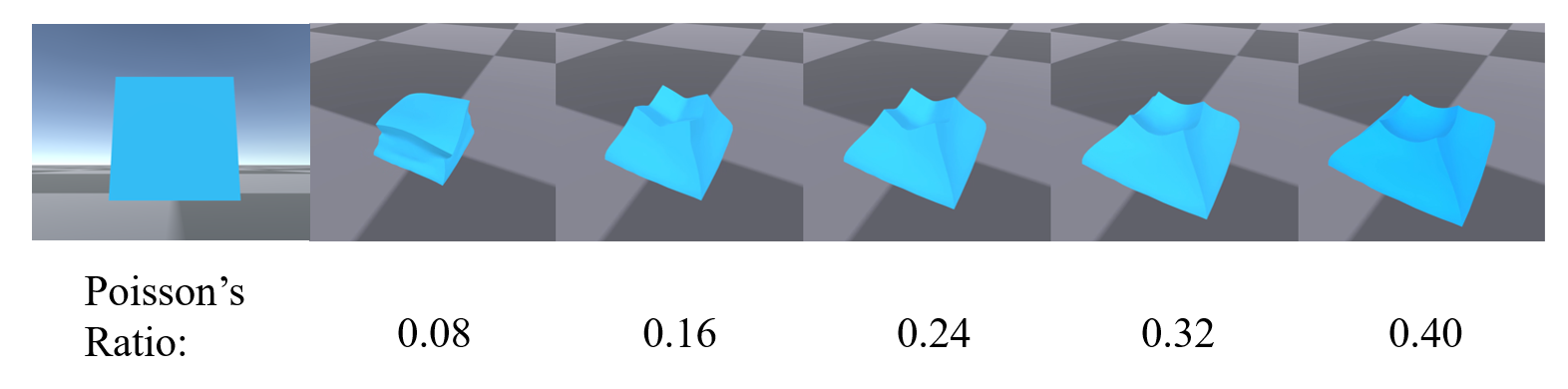}
    \caption{Soft cubes with different Poisson's ratio falling on the ground. Here we choose Young's modulus to be 1e3. }
    \label{fig:ablation-poissons-ratio}
    \vspace{-0.3cm}
\end{figure}

From Fig. \ref{fig:ablation-youngs-modulus}, we can observe that as Young's modulus increases, the soft cube becomes more rigid and stiff, and the garment generates more wrinkles since it becomes more difficult for the garment faces to stretch. As shown in \ref{fig:ablation-poissons-ratio}, when the Poisson's ratio increases, the soft cube surface becomes more concave, making it easier for the soft cube elements to experience shearing deformation. 
We also test the shearing stiffness of the garment material. The results are not presented in the paper but are included in the supplementary material on the project website, as their differences are tiny. In all of these tests, the soft cube uses the neo-Hookean model, and the garment uses the Witkin-Baraff model. 

\begin{figure}
\vspace{0.2cm}
    \centering
    \includegraphics[width=1\linewidth]{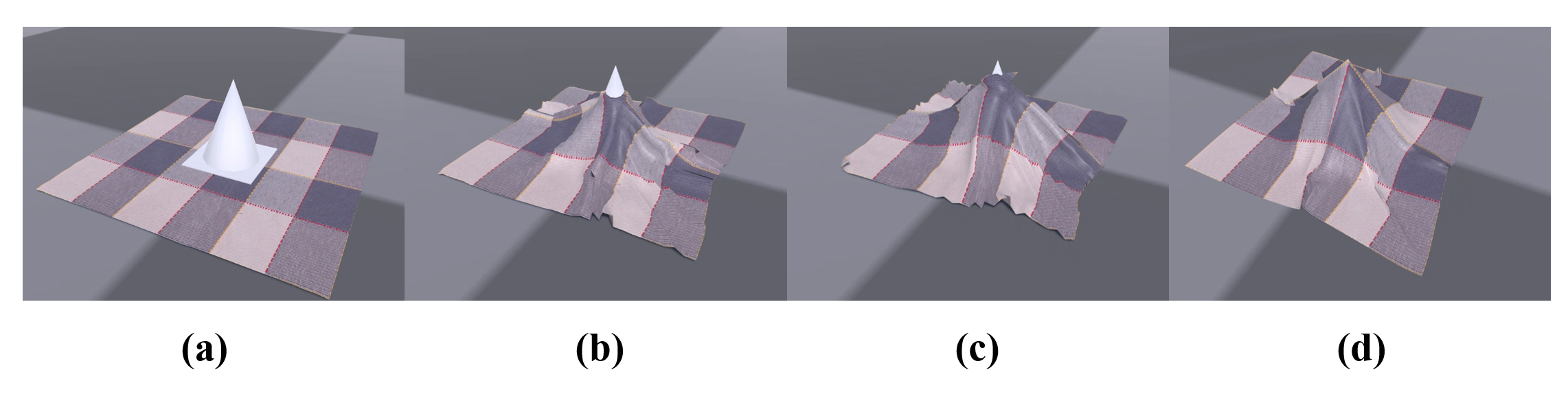}
    \caption{When a garment falls onto rigid cones using varied soft-rigid coupling methods, distinct outcomes arise. In (a), the absence of coupling allows the garment to pass through the cone. In (b), impulse-based coupling hinders passage but leads to significant mesh penetrations. Despite using a surface collision detection in (c), impulse-based coupling still can't prevent mesh intersections. Conversely, (d) ZeMa achieves intersection-free and realistic contact results.}
    \label{fig:efficacy_of_coupling}
    \vspace{-0.3cm}
\end{figure}

\paragraph{Efficacy of Coupling}
The coupling between soft and rigid bodies is necessary to gain reasonable contact effects. We use the scene of a garment falling onto a rigid cone as an example. As shown in Fig. \ref{fig:efficacy_of_coupling}, without any such coupling handling, the garment will completely penetrate the rigid cone and subsequently lie on the ground. With the constraints-based weak coupling and a particle-based collision detection scheme, the garment will still penetrate the cone and become stuck in the cone. The intersections between the garment and the cone cannot be eliminated, even enabling surface collision detection. To note, these two weak coupling cases are common implementations in current popular simulators. ZeMa, benefitting from the barrier-based soft-rigid coupling, generates intersection-free simulation results with reasonable collision effects, proving its coupling efficacy.  

\paragraph{Runtime over different DoFs}

\begin{table}[ht!]
    \centering
    \begin{tabular}{|c|ccccc|}
    \hline
       \#Vertices  & 680 & 2083 & 11094  & 19049 & 77995 \\ \hline
       \#Triangles & 1742 & 4816 & 24565  & 41631 & 160703 \\ \hline
       Runtime (s) & 3.23 & 3.30 & 3.58 &  3.48 & 15.21\\ \hline
    \end{tabular}
    \caption{Simulation runtime in scenes with different numbers of geometric elements. }
    \label{tab:simulation_runtime}
    \vspace{-0.2cm}
\end{table}

We employ an object mesh comprising 160K triangles and 78K vertices, which is then re-meshed using the quadric edge collapse decimation algorithm \cite{simplification} to yield five meshes of varying resolutions, including the original. These mesh-based rigid objects are utilized to assess the ZeMa runtime, with outcomes presented in Table \ref{tab:simulation_runtime}. The table indicates that computational complexity does not scale linearly with increasing vertex and triangle counts but rather at a rate significantly slower than mesh resolution. This is attributed to the system's DOF being influenced by the number of rigid bodies, not their individual vertices or triangles, facilitating precise and efficient simulations of intricate geometries. Please refer to supplementary materials for soft object runtime details.

\paragraph{Accuracy Evaluation}

To evaluate the sim-to-real gap of ZeMa, we executed two experiments using a gel elastomer with markers, both in reality and within ZeMa. In the first, a robot gripper, with an indenter affixed, grasps a cube and depresses the gel on a table by 0.5mm before rotating 0.3 rad vertically. We monitor the movement of markers on the elastomer using a camera positioned below and simulate the scenario in ZeMa for comparison. The subsequent experiment mirrors the first, but the gripper shifts 1 mm horizontally post-press instead of rotating. A calibrated camera captures real-world marker displacements, and an analogous camera in ZeMa captures simulated displacements. Tracking these markers, ZeMa's predicted errors were 0.78px and 1.94px, with relative errors of 5.1\% and 5.5\%, underscoring ZeMa's alignment with real-world outcomes. A detailed experimental setup is available in the supplementary materials.

\section{Conclusion and Future Works}
\label{sec:conclusion}
In this work, we introduce ZeMa, an advanced simulation platform tailored for robotic manipulation involving soft objects. Addressing a notable gap in the field, ZeMa proficiently combines the requisite attributes of two-way soft-rigid coupling, intersection-free dynamics, and frictional contact modeling, all while maintaining a competitive runtime suitable for deep and reinforcement learning. Not only does it have physical accuracy and surpass the runtime of the baseline method such as IPC-GraspSim by a remarkable 75-fold, but its practical utility is also evidenced through successful applications in parallel grasp generation, penetrated grasp repair, and real-world reinforcement learning for grasping. In the future, we hope our work can serve as a foundational tool for research in deformable object manipulation or soft robotics.

\bibliographystyle{IEEEtran}
\bibliography{zema}

\end{document}